\documentclass[10pt,twocolumn,letterpaper]{article}

\usepackage[pagenumbers]{cvpr} 
\usepackage{bm}
\usepackage{amsmath}
\usepackage{amssymb}
\usepackage{xcolor}
\definecolor{Graylight}{gray}{0.9}
\definecolor{Gray}{gray}{1.0}
\usepackage{colortbl}
\usepackage{booktabs}
\usepackage{multirow}
\usepackage{mathrsfs}
\usepackage{marvosym}
\usepackage{enumerate}
\usepackage[hang,flushmargin]{footmisc} 
\usepackage{tablefootnote}

\definecolor{cvprblue}{rgb}{0.21,0.49,0.74}
\usepackage[pagebackref,breaklinks,colorlinks,allcolors=cvprblue]{hyperref}
\usepackage{multirow}

\def\paperID{*****} 
\def\confName{CVPR}
\def\confYear{2025}

\title{InternLM-XComposer2.5-OmniLive: A Comprehensive Multimodal System for Long-term Streaming Video and Audio Interactions }

\author{Pan Zhang$^{*1}$, Xiaoyi Dong$^{*1,2}$, Yuhang Cao$^{*1}$, Yuhang Zang$^{*1}$, Rui Qian$^{*1,2 \dag}$, Xilin Wei$^{1,3 \dag}$, Lin Chen$^{1,4 \dag}$, \\ Yifei Li${^{1,5 \dag}}$, Junbo Niu$^{1,6 \dag}$, Shuangrui Ding$^{1,2 \dag}$, Qipeng Guo$^{1}$, Haodong Duan$^{1}$, Xin Chen$^{1}$, Han Lv$^{1}$, \\ Zheng Nie$^{1}$, Min Zhang$^{1}$, Bin Wang$^{1}$, Wenwei Zhang$^{1}$, Xinyue Zhang$^{1}$, Jiaye Ge$^{1}$, Wei Li$^{1}$, Jingwen Li$^{1}$, \\ Zhongying Tu$^{1}$,  Conghui He$^{7}$, Xingcheng Zhang$^{7}$, Kai Chen$^{1}$, Yu Qiao$^{1}$, Dahua Lin$^{1,2}$, Jiaqi Wang$^{1,}${\textsuperscript{\Letter}}\\
$^1$Shanghai Artificial Intelligence Laboratory,  $^2$The Chinese University of Hong Kong, \\ $^3$ Fudan University, $^4$ University of Science and Technology of China,  \\ $^5$ Tsinghua University, $^6$ Beihang University, $^7$ SenseTime Group \\
{\tt\small internlm@pjlab.org.cn}
}

\begin{document}
\maketitle

{\let\thefootnote\relax\footnotetext{\noindent* indicates equal contribution. $\dag$ indicates interns at IXCLab, Shanghai AI Laboratory}}

\begin{abstract}
Creating AI systems that can interact with environments over long periods, similar to human cognition, has been a longstanding research goal. Recent advancements in multimodal large language models (MLLMs) have made significant strides in open-world understanding. However, the challenge of continuous and simultaneous streaming perception, memory, and reasoning remains largely unexplored. Current MLLMs are constrained by their sequence-to-sequence architecture, which limits their ability to process inputs and generate responses simultaneously, akin to being unable to think while perceiving. Furthermore, relying on long contexts to store historical data is impractical for long-term interactions, as retaining all information becomes costly and inefficient. Therefore, rather than relying on a single foundation model to perform all functions, this project draws inspiration from the concept of the \textbf{Specialized Generalist AI} and introduces disentangled streaming perception, reasoning, and memory mechanisms, enabling real-time interaction with streaming video and audio input. The proposed framework \textbf{InternLM-XComposer2.5-OmniLive (IXC2.5-OL)} consists of three key modules: \textbf{(1) Streaming Perception Module}: Processes multimodal information in real-time, storing key details in memory and triggering reasoning in response to user queries. \textbf{(2) Multi-modal Long Memory Module:} Integrates short-term and long-term memory, compressing short-term memories into long-term ones for efficient retrieval and improved accuracy. \textbf{(3) Reasoning Module:} Responds to queries and executes reasoning tasks, coordinating with the perception and memory modules. This project simulates human-like cognition, enabling multimodal large language models to provide continuous and adaptive service over time. All code and models of \textbf{InternLM-XComposer2.5-OmniLive (IXC2.5-OL)} are publicly available at \url{https://github.com/InternLM/InternLM-XComposer/tree/main/InternLM-XComposer-2.5-OmniLive}.
\end{abstract}
\section{Introduction}
\label{sec:intro}

\begin{figure}[t!]
    \centering
    \includegraphics[width=1.0\linewidth]{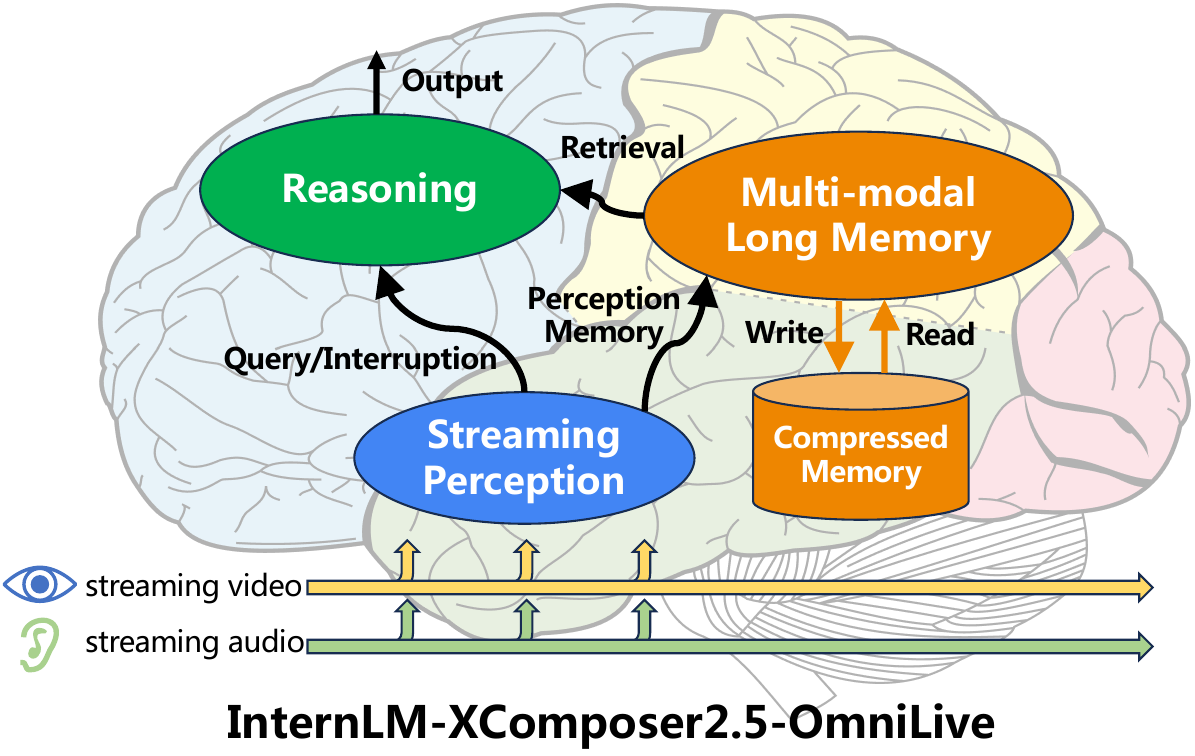}
    \setlength{\abovecaptionskip}{0mm} 
    \captionof{figure}{\small Inspired by human-like cognition and Specialized Generalist AI, we introduce InternLM-XComposer2.5-OmniLive (IXC2.5-OL), a system that facilitates real-time interaction with: (1) a \textbf{streaming perception} module supports streaming video and audio inputs; (2) a \textbf{multi-modal long memory} module that compresses short-term memory into long-term memory; and (3) a \textbf{reasoning} module that answers queries based on retrieved memories.}
    \label{fig:teaser}
\end{figure}

The goal of developing AI systems~\cite{lecun2022path} that can understand and interact with environments over long periods, akin to human cognition, has been a central focus of research for decades. The rise of large-scale data corpora~\cite{lin2014microsoft,kuznetsova2020open,schuhmann2022laion, wang2023v3det} and multimodal large language models~\cite{openai2023gpt4vision,hurst2024gpt,geminiteam2023gemini} has driven significant advances in free-form multimodal question answering. Recent developments, such as Mini-Omni~\cite{xie2024mini}, VideoLLM-Online~\cite{chen2024videollmonlineonlinevideolarge}, and VITA~\cite{fu2024vitaopensourceinteractiveomni}, have made notable strides toward enabling more natural and immersive online interactions. However, challenges persist in creating systems capable of continuous interaction due to the intrinsic limitations of a single decoder-only large language model architecture.

Existing architectures~\cite{zhang2024longva,xie2024mini,chen2024videollmonlineonlinevideolarge,fu2024vitaopensourceinteractiveomni} encounter significant limitations in real-time and long-term streaming perception, reasoning, and memory. The sequence-to-sequence decoder-only architecture used in current MLLMs forces a switch between perception (e.g., seeing and hearing) and thinking, limiting the simultaneous processing of inputs and outputs. Additionally, existing works~\cite{zhang2024flashvstreammemorybasedrealtimeunderstanding,wang2024videollamblongcontextvideounderstanding,fan2024videoagentmemoryaugmentedmultimodalagent} rely on the integration of multimodal memories within context windows. The reliance on long contexts to store historical information proves impractical for long-term use, especially in scenarios requiring continuous AI assistance. Multimodal data, like video streams, can quickly accumulate millions of tokens within a few hours, making it impractical to maintain context over multiple days of service. The cost and inefficiency of storing all historical clues within the context further limit the system's capacity to provide continuous and long-term service. In contrast, the human brain can effortlessly integrate perception and cognition, preserving long-term multimodal memories. This is believed to be closely related to the functional partitioning design of the human brain cortex, where different areas of the cortex are responsible for distinct tasks, such as perception, memory, and cognition.

Inspired by the paradigm of Specialized Generalist AI~\cite{zhang2024towards}, we propose a system \textbf{InternLM-XComposer2.5-OmniLive (IXC2.5-OL)} composed of fused specialized generalist models for streaming perception, reasoning, and memory, respectively. The system is designed to enable AI models to engage continuously with environments while retaining observations over time. By integrating short-term and long-term multimodal memory, our approach attempts to emulate human-like cognition, enabling more dynamic and sustained interactions.

As shown in Figure~\ref{fig:teaser}, the IXC2.5-OL system consists of three key modules: (1) \textbf{Streaming Perception Module:} This module processes the multimodal information stream on-the-fly. To ensure perception accuracy and efficiency, the video and audio streams are handled separately. A live video perception model processes the video stream, encoding the information and storing key details in memory. Meanwhile, an audio model recognizes the contents of human speech and other sounds, \eg, barking, knocking, or whistling. It triggers the reasoning process when human queries occur.
(2)  \textbf{Multi-modal Long Memory Module}: This component integrates both long-term and short-term memory, enabling the retrieval of detailed short-term information as well as long-term historical cues. It continuously compresses short-term memories into more information-rich long-term memories to enhance retrieval efficiency and accuracy.
(3)  \textbf{Reasoning Module}: The reasoning module, activated by the perception module, handles queries and performs reasoning tasks. As the component with the most model parameters, it serves as the core of the system's deep cognitive processes.

The proposed system empowers AI with the ability to perceive, think, and memorize simultaneously. By overcoming the limitations of alternating perception and reasoning, IXC2.5-OL seeks to provide continuous, adaptive service, and long-term AI service. The proposed system will not only enhance the performance of AI assistants but will also contribute to the broader AI applications capable of continuously interacting and adapting to dynamic environments.

The \textbf{IXC2.5-OL} demonstrates strong performance across both audio and video benchmarks. Among the open-source models, IXC2.5-OL achieves competitive results on audio recognition (ASR) benchmarks such as Wenetspeech~\cite{zhang2022wenetspeech} for Chinese and LibriSpeech~\cite{panayotov2015librispeech} for English.
For video understanding benchmarks, IXC2.5-OL achieves state-of-the-art results among models with less than 10B parameters, obtaining an M-Avg of 66.2\% on MLVU \cite{zhou2024mlvu} and an overall accuracy of 68.7\% on MVBench \cite{li2024mvbench}. Additionally, it demonstrates competitive performance on Video-MME \cite{fu2024video} (60.6\%) and MMBench-Video \cite{fang2024mmbench} (1.42).
On recent streaming video bench StreamingBench~\cite{lin2024streamingbench}, IXC2.5-OL achieves new SOTA results on open-source models (73.79\%), highlighting its exceptional capabilities for real-time video interactions.

To foster the development of the multimodal streaming interaction community, alongside the model parameters, the inference and deployment source code, encompassing both the web frontend and backend code, has also been released. All code and models of IXC2.5-OL are publicly available at \url{https://github.com/InternLM/InternLM-XComposer/tree/main/InternLM-XComposer-2.5-OmniLive}.

\section{Related Works}
\label{sec:rel}
\noindent \textbf{MLLMs for Text-Image Conversation.} Large Language Models (LLMs)~\cite{brown2020language,ouyang2022training,openai2020chatgpt,chowdhery2022palm,kaplan2020scaling,touvron2023llama,touvron2023llama2,jiang2023mistral,2023internlm,zeng2023glm-130b,baichuan2023baichuan2,qwen7b,cai2024internlm2} have garnered significant attention for their remarkable capabilities in language comprehension and generation. Building on this success, Large Vision-Language Models (LVLMs)~\cite{zhang2023internlm,openai2023gpt4,chen2023pali,chen2023palix,chen2023pali3,driess2023palme,fu2023gemini,zhu2023minigpt,dai2023instructblip,zhang2023internlm,fuyu-8b,li2023otter,peng2023kosmos,ye2023mplug,awadalla2023openflamingo,internlmxcomposer2_4khd,lin2024vilapretrainingvisuallanguage} have been developed by integrating LLMs with vision encoders~\cite{radford2021learning,zhang2024long,sun2023alpha,zhai2023sigmoid,oquab2024dinov2learningrobustvisual,zang2023contextual,liu2022convnet2020s,chen2023internvl,chen2023sharegpt4v,lin2023sphinx,bai2023qwen,wang2023cogvlm,internlmxcomposer2,cao2024dualfocus,liu2024rar,chen2024fargpt4vclosinggap,zhang2024omgllavabridgingimagelevelobjectlevel}, extending their ability to comprehend visual content and enabling applications like text-image conversations. Earlier LVLMs were primarily designed for single-image, multi-round conversations, whereas recent advancements~\cite{alayrac2022flamingo,bai2023qwen,zhao2023mmicl,sun2024generativemultimodalmodelsincontext,lin2024vilapretrainingvisuallanguage,jiang2024mantisinterleavedmultiimageinstruction,internlmxcomposer2_4khd, internlmxcomposer2_5,li2024ariaopenmultimodalnative} have expanded their capabilities to process and understand multi-image inputs.

\noindent \textbf{MLLMs for Video Understanding.} In addition to advancements in image understanding, the field of MLLMs has seen growing efforts in video analysis~\cite{li2023mvbench,ning2023video,liu2024tempcompass,caba2015activitynet,fang2024mmbench,song2023moviechat,song2024moviechat+,xue2024longvilascalinglongcontextvisual,wang2024qwen2vlenhancingvisionlanguagemodels}. To address the complexity of video inputs, existing approaches leverage techniques such as sparse sampling or temporal pooling~\cite{lin2023video,maaz2023video,luo2023valley,huang2024image,yu2024self}, compressed video tokens~\cite{li2023videochat,zhang2023video,jin2023chat,weng2024longvlm,li2023llama,ryoo2024xgenmmvid,chen2024timemarkerversatilevideollmlong}, and memory banks~\cite{song2023moviechat,he2024ma,song2024moviechat+,qian2024streaminglongvideounderstanding,zhang2024flashvstreammemorybasedrealtimeunderstanding,wang2024videollamblongcontextvideounderstanding,fan2024videoagentmemoryaugmentedmultimodalagent}. Additionally, some methods utilize language as a bridge for video understanding~\cite{kahatapitiya2024language,islam2024video,zhang2023simple}. Beyond these video-specific strategies, video analysis can also be framed as interpreting a high-resolution composite image generated from sampled video frames~\cite{kim2024imagegridworthvideo,xu2024pllavaparameterfreellava,zhang2024longva}. Recent advancements~\cite{chen2024videollmonlineonlinevideolarge,wang2024videollmknowsspeakenhancing,wu2024videollmmodefficientvideolanguagestreaming,zhang2024flashvstreammemorybasedrealtimeunderstanding} have increasingly focused on online video understanding, aiming to simulate real-world scenarios where AI processes video streams in real-time to comprehend the environment on-the-fly. However, existing solutions still lack the capability to simultaneously perform perception, memory, and reasoning, limiting their applicability for consistent and long-term human-AI interactions.

\noindent \textbf{MLLMs for Audio Understanding.} Audio understanding can be effectively modeled as a sequence-to-sequence (Seq2Seq) task~\cite{radford2023robust}, which enables powerful integration with large language models by incorporating audio tokenizers and encoders~\cite{tang2023salmonn,zhang2023speechgpt,chu2023qwen,zeng2024glm}.
In addition to receiving the audio input, recent research investigates streaming duplex speech models~\cite{wang2024full,yu2024salmonn,ma2024language,wang2024freeze} that allow speakers to interrupt freely.
Beyond audio-text models, emerging research delves into audio-visual models ~\cite{shu2023audio,li2024saven} and unified architectures that process audio, visual, and text modalities~\cite{zhan2024anygptunifiedmultimodalllm,fu2024vitaopensourceinteractiveomni,li2024oceanomniunderstandworldomnimodality}.

\noindent \textbf{MLLMs for Omni-Modal Understanding.} Integrating multiple modalities into a single omni-modal foundation model represents a promising research direction. Existing works~\cite{han2023onellmframeworkalignmodalities,zhan2024anygptunifiedmultimodalllm,wu2024nextgptanytoanymultimodalllm,li2024oceanomniunderstandworldomnimodality,fu2024vitaopensourceinteractiveomni,chen2024emovaempoweringlanguagemodels,xie2024miniomni2opensourcegpt4ovision,sun2024videosalmonnspeechenhancedaudiovisuallarge} explore models capable of processing omni-modal inputs, typically combining video and audio, to produce outputs in various formats. These outputs include text~\cite{han2023onellmframeworkalignmodalities,li2024oceanomniunderstandworldomnimodality,fu2024vitaopensourceinteractiveomni}, audio~\cite{chen2024emovaempoweringlanguagemodels,sun2024videosalmonnspeechenhancedaudiovisuallarge,xie2024miniomni2opensourcegpt4ovision}, and omni-modal contents~\cite{zhan2024anygptunifiedmultimodalllm,wu2024nextgptanytoanymultimodalllm}. In the current design of IXC2.5-OL, we handle the audio and video modalities separately to mitigate potential influence during joint training. In future versions, our model will incorporate joint training across all modalities, enabling seamless omni-modality integration.
\section{Method}

\begin{figure*}
    \centering
    \includegraphics[width=0.9\linewidth]{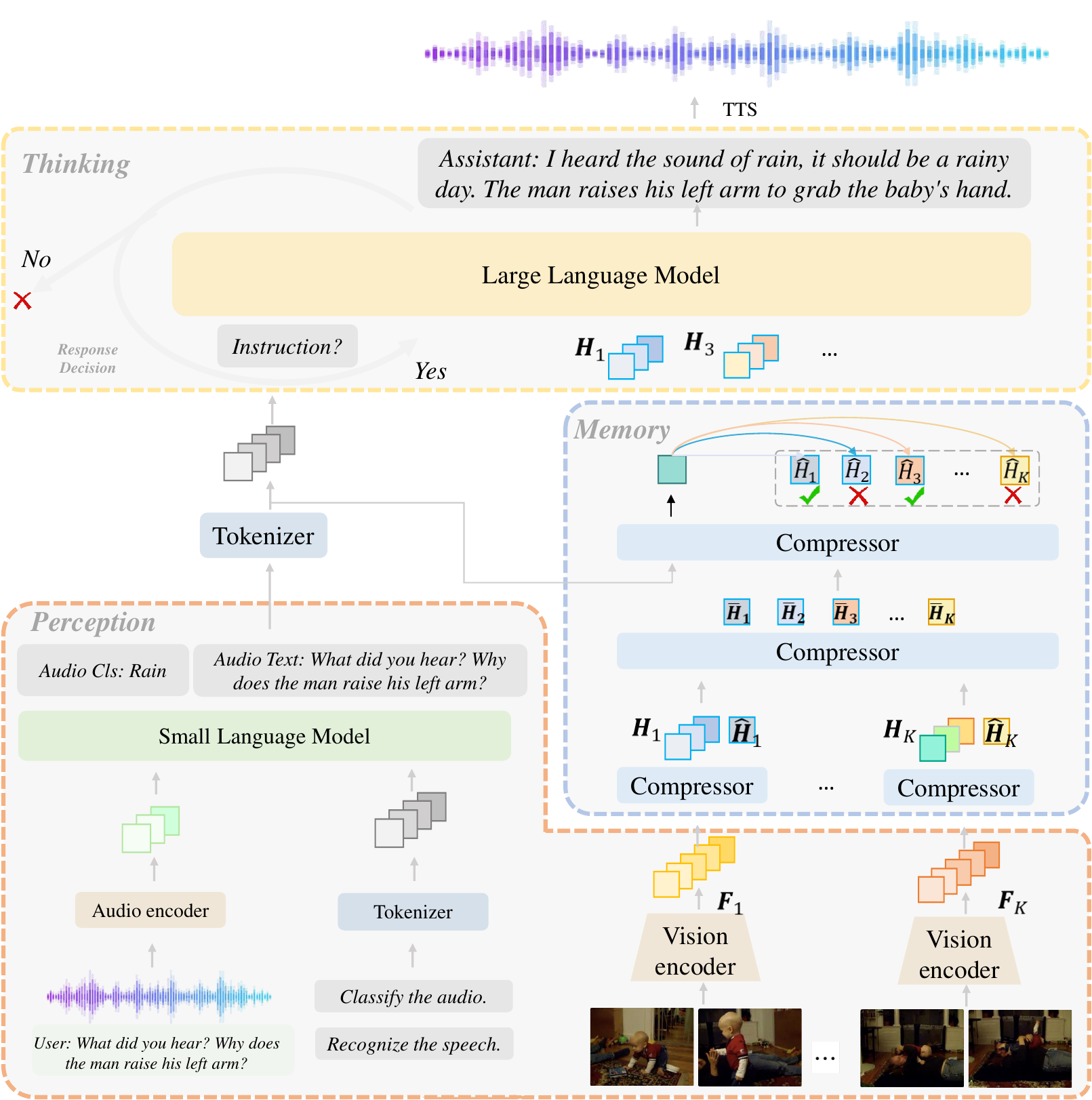}
    \caption{\textbf{Pipeline of the InternLM-XComposer2.5-OmniLive. (IXC2.5-OL)}. The IXC2.5-OL is a real-time interacting system that is constructed by three simultaneous modules: 1) the Streaming Perception Module, 2) the Multi-modal Long Memory Module, and 3) the Reasoning Module.}
\end{figure*}

As we briefly introduced in Sec.1, the IXC2.5-OL has three disentangled modules: 1) the Streaming Perception Module for on-the-fly visual and audio information processing, 2) the Multi-modal Long Memory Module for memory integration and retrieval, and 3) the Reasoning Module collect information from the perception and memory module, and handles queries and performs reasoning tasks. All the modules work simultaneously and interact asynchronously. 

\subsection{Streaming Perception Module}
Besides nature language, the IXC2.5-OL could handle video and audio natively. To realize this, the Streaming Perception Module contains an Audio Translation Module and a Video Perception Module.

\begin{table}[t]
\centering
\caption {Overview of datasets used in pretraining and supervised fine-tuning (SFT) for the Audio Translation Module. The pretraining stage focuses solely on the automatic speech recognition (ASR) task, utilizing the GigaSpeech and WenetSpeech datasets. The SFT stage includes both ASR and audio classification (CLS) tasks, leveraging diverse datasets. For CommonVoice, we only use the English and Chinese splits. Additionally, 475 self-constructed “Silence” samples are used for CLS tasks.}
\small
\label{tab:audio-data} 
\begin{tabular}{lccc}
\toprule
\textbf{Stage} & \textbf{Task}   & \textbf{Dataset} & \textbf{Data Num} \\ 
\midrule
\multirow{2}{*}{Pretrain} & \multirow{2}{*}{ASR} & GigaSpeech \cite{chen2021gigaspeech} & 8,282,987 \\
\multirow{15}{*}{SFT} & \multirow{12}{*}{ASR} & WenetSpeech \cite{zhang2022wenetspeech} & 17,821,017 \\ \midrule
      &    & LibriSpeech \cite{panayotov2015librispeech} & 281,241 \\
      &    & VCTK     \cite{Veaux2017CSTRVC}  & 44,070  \\
      &    & AISHELL-1 \cite{bu2017aishell}  & 120,098 \\ 
      &    & AISHELL-4 \cite{fu2021aishell}  & 102,254 \\ 
      &    & MD-RAMC \cite{yang2022open}  & 219,325 \\ 
      &    & ASCEND \cite{lovenia2021ascend}  & 12,314 \\ 
      &    & KeSpeech \cite{tang2021kespeech}  & 888,428 \\ 
      &    & DASR \cite{cornell2024chime}  & 190,732 \\ 
      &    & CommonVoice \cite{ardila2019common}  & 2,813,852 \\
      \cmidrule(r){2-4}
      & \multirow{3}{*}{CLS}   & FSD50K \cite{fonseca2020fsd50k}  & 40,966 \\ 
      &    & AudioSet \cite{kong2018audio}  & 18,683 \\ 
      &    & Silence    & 475 \\ 
\bottomrule
\end{tabular}
\end{table}

\noindent\textbf{Audio Translation Module} contains an audio encoder, an audio projector, and a Small Language Model (SLM). The audio encoder encodes the input audio sample into high-dimension features, and the audio projector further maps the feature to the input space of the SLM. The SLM outputs both the class (e.g. laughing, clapping, or raining) of the audio and the natural language within the audio (i.e. the automatic speech recognition). 
In practice, we use the Whisper~\cite{radford2022robustspeechrecognitionlargescale} model as the audio encoder and a Qwen2-1.8B~\cite{yang2024qwen2} as the SLM. The training contains two stages and we list the training data in Table \ref{tab:audio-data}.

\noindent\textbf{Video Perception Module} provides coarse-grained visual information to the Multi-modal Long Memory Module. It processes the real-time video input stream and encodes each frame into semantic features. For efficiency, we use the OpenAI CLIP-L/14~\cite{radford2021learning} In practice.

\subsection{Multi-modal Long Memory Module}
The Multi-modal Long Memory Module is the core design to handle extremely long video input and helps the Reasoning Module to get rid of millions of tokens from its context window. It shares a similar idea from the VideoStreaming~\cite{qian2024streaminglongvideounderstanding} that encodes video clips into short-term memories and integrates them into long-term memory. With the given questions, it retrieved the most related video clips for the Reasoning Module. Formally, the Multi-modal Long Memory Module is trained with three tasks:

\noindent\textbf{Video Clip Compression.} With features of $k_{th}$ video clip extracted from the Perception Module $\bm{F}_k\in\mathbb{R}^{TN\times C}$, we initialize its short-term memory $\bm{H}_k\in\mathbb{R}^{TP\times C} $ by the spatial down-sampling and its global memory $\hat{\bm{H}_k}\in\mathbb{R}^{1\times C}$. We realize the compression by the auto-regressive and feature aggregation nature of LLMs:
\begin{align*}
    \bm{H}_k,\hat{\bm{H}}_k = Compressor([\bm{F}_k\circ\bm{H}_k\circ\hat{\bm{H}_k]}).
\label{eq_compress}
\end{align*}

\noindent\textbf{Memory Integration.} 
Short-term memory represents the detailed information of each short video clip while the model still lacks a macro view of the video. To this end, with the short-term and global memory of a list of video clips, we integrate them into long-term memory by the Compressor in the following format:
 
\begin{align*}
    & \bar{\bm{H}}_1, \bar{\bm{H}}_2, ..., \bar{\bm{H}}_k = \\ & Compressor([\bm{H}_1\circ\bm{H}_2...\circ\bm{H}_k\circ\hat{\bm{H}_1}\circ\hat{\bm{H}_2}...\circ\hat{\bm{H}_k}]). 
\end{align*} 

the $\bar{\bm{H}} = [\bar{\bm{H}}_1, \bar{\bm{H}}_2, ..., \bar{\bm{H}}_k] \in\mathbb{R}^{k\times C}$ represents the video in a high-compressed way and we denote it as the long-term memory.

\noindent\textbf{Video Clip Retrieval.} 
When users raise questions, the Multi-modal Long Memory Module retrieves the question-related video clips and provides both the video clips and their short-term memory to the Reasoning Module. In practice, we first encode the question to the feature space of the memory. We concatenate the long-term memory with the tokenized question as the Compressor input, and we view the last token of the output features as the memory-space-aligned question feature.
Then we calculate the similarity between the question feature and each video's global memory, and select the most related clips for the Reasoning Module. 

\noindent\textbf{Implementation Detail.} 
We use Qwen2-1.8B~\cite{yang2024qwen2} as the LLMs and construct several kinds of training data for the three aforementioned tasks. As shown in Table.~\ref{tab:sft data}, we train the Video Clip Compression task with short video captioning data from multiple sources, using the same prefix captioning task designed in VideoStreaming~\cite{qian2024streaminglongvideounderstanding}. For the Memory Integration task and Video Clip Retrieval task, besides the off-the-shelf video grounding data, we also construct data for two unique tasks: `Semantics Implicit Question' and `Reference Implicit Question'. 

The `Semantics Implicit Question' means the question does not point to some object directly, but mentions the usage or meaning of the object, and the model should find out the object by understanding the implicit question. For example, when the user asks `How about the weather today?', the model should find out some weather-related object in the past video stream, such as an umbrella, a sun-glass, or something. Another example could be `I'm hungry, where can I heat my sandwiches?', the model should find the microwave oven it has seen before. 

The `Reference Implicit Question' means the question uses pronouns rather than nouns. For example, `What is this' means the models should retrieve the current frames, although it does not mention any exact objects. 

Both kinds of implicit questions are commonly used in real-world communication while current models failed to handle them, so we construct corresponding training data to empower the model with these capabilities.

\begin{table}[t]
\centering
\footnotesize
\setlength{\tabcolsep}{3mm}{
\begin{tabular}{ll}
\toprule
Model &  Dataset\\
\midrule 
\multirow{4}{*}{Memory Module} & ShareGPT4Video \cite{chen2024sharegpt4video}, Ego4D\cite{grauman2022ego4d} \\
 & ActivityNet~\cite{caba2015activitynet} \\
& Semantics Implicit QA \\ 
& Reference Implicit QA \\ \midrule
\multirow{3}{*}{IXC2.5} & ShareGPT4Video \cite{chen2024sharegpt4video}, ActivityNet~\cite{caba2015activitynet} \\
 & FunQA~\cite{xie2025funqa},  TrafficQA~\cite{trafficqa} \\
 & VideoChat2-IT\cite{li2023mvbench},  LLaVA-Video~\cite{llavavideo} \\  

\bottomrule
\end{tabular}}
\vspace{-6pt}
\caption {\textbf{Video Datasets used in IXC2.5-OL}. }
\label{tab:sft data}
\vspace{-12pt}
\end{table}

\subsection{Reasoning Module}
The Reasoning Module is initialized by an improved version of InternLM-XComposer2.5 (IXC2.5 in the following for simplified statement) and we add a memory projector to align the memory feature with IXC-2.5. 
For a given questions and both visual and memory information provided by the Memory Module, we formulate the input as:
\begin{align*}
    & \textrm{Question:} <|\textrm{Que}|>,\\
    & \textrm{Here is the question related video clip} <|\textrm{Img}|>;  \\
    & \textrm{Here is the question related memory} <|\textrm{Mem}|>  \\    
\end{align*}

In real-world usage, there exists some noisy input that should not be answered (e.g., the user says `enn...' or `ok...'), the model should keep salient and wait for the next question. To realize this, we add an additional `Instruction Prediction' process for each question to decide it should be answered or not.

\begin{figure*}
    \centering
    \includegraphics[width=0.98\linewidth]{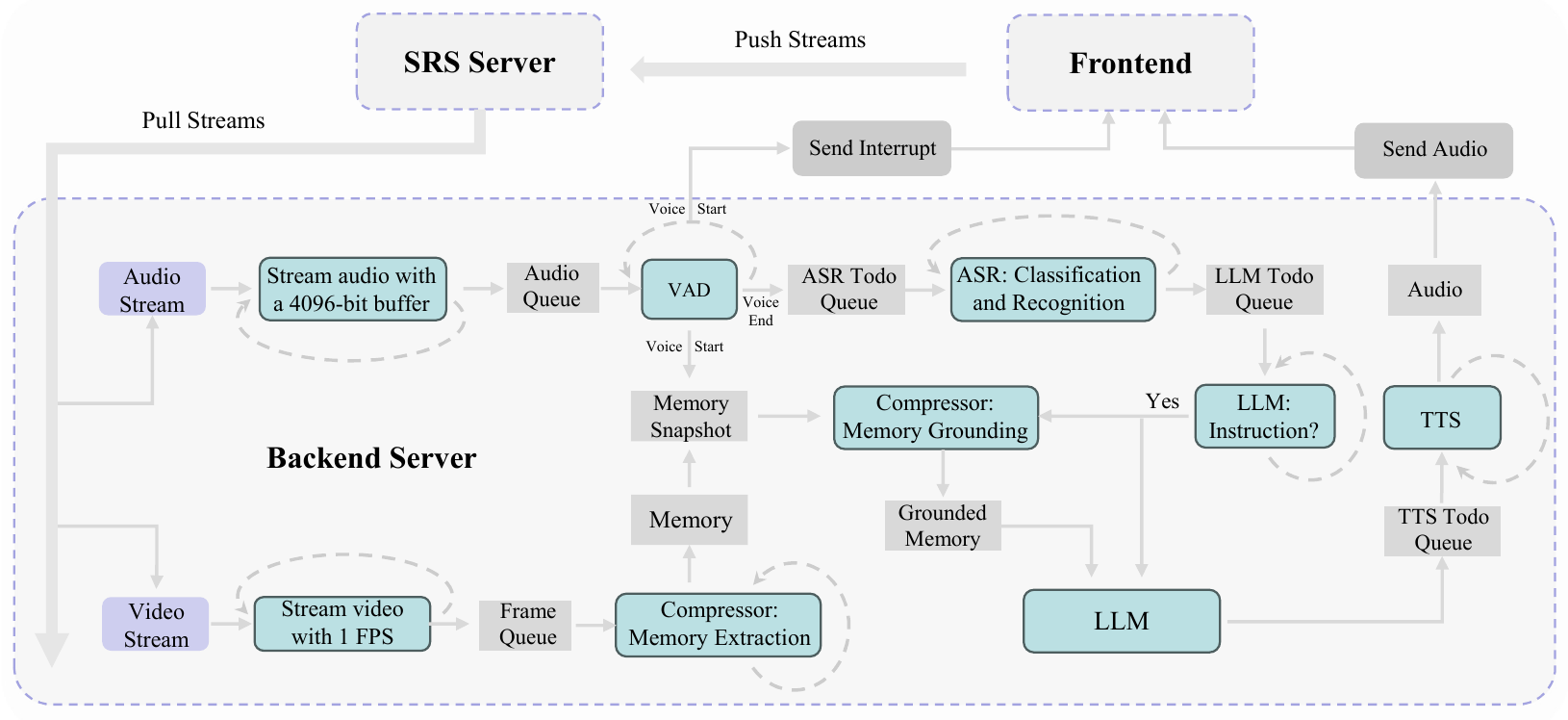}
    \caption{\textbf{System pipeline of the IXC2.5-OL}. The system comprises the Frontend, SRS Server, and Backend Server. The Frontend is utilized for capturing video and audio streams and for playing audio from the Backend Server. The SRS Server is employed for managing live streams. The Backend Server is responsible for reading audio and video, extracting memory, and answering questions. The green boxes in the figure represent a thread or a process.}
    \label{fig:system}
\end{figure*}

\subsection{System Pipeline}
As illustrated in Figure \ref{fig:system}, the system comprises the Frontend, SRS Server, and Backend Server.

\noindent\textbf{Frontend.} 
The frontend application, developed with JavaScript, enables the camera and microphone to capture video and audio stream inputs, which are then pushed to the SRS server. Concurrently, it establishes a WebSocket connection with the backend to listen for audio outputs and interrupt signals. When audio data is received, the frontend plays it. Upon receiving an interrupt signal, the frontend suspends the audio playback and discards the pending audio.

\noindent\textbf{SRS Server.} 
SRS (Simple Realtime Server) is a straightforward and efficient real-time video server, adept at supporting a multitude of real-time streaming protocols such as RTMP, WebRTC, HLS, HTTP-FLV, SRT, and others. It is renowned for its ability to reliably receive and deliver audio and video streams.

\noindent\textbf{Backend Server.} 
After establishing a WebSocket connection with the frontend, the backend will pull streaming from the SRS Server and initiate separate threads to read audio and video. 

The audio reading thread will segment the audio stream into 4096-bit chunks and enqueue them into the \emph{Audio Queue}. The Voice Activity Detection (VAD)~\cite{gao2023funasr} thread continuously reads data from \emph{Audio Queue} and detects the start and end of voice activity. Upon detecting the start of voice activity, the backend sends an interrupt signal to the frontend to pause the currently playing audio, and at the same time, dispatches a backup signal to the video process, directing it to save the current memory state. When detecting the end of voice activity, the entire voice segment will be enqueued into \emph{ASR Todo Queue}. The ASR thread continuously reads audio segments from \emph{ASR Todo Queue}, performs background noise classification and voice recognition on them, and then enqueues the results into \emph{LLM Todo Queue} for use by the LLM.

The video reading thread reads video frames at a rate of 1 frame per second and enqueues them into \emph{Frame Queue}. The compressor process reads video frames from the queue, recognizes them, extracts relevant memory, and stores it. Upon receiving a backup signal from the VAD thread, the compressor process will save the current memory state for later retrieval.

The LLM process reads text from the \emph{LLM Todo Queue} and determines whether it is an instruction that requires a response from the model. For texts identified as instructions, the compressor process will use the current instruction and the backed-up memory to perform memory grounding, in order to retrieve memories related to the instruction. The LLM process will then generate a response based on the retrieved memories and the instruction, and enqueue the resulting output into \emph{TTS Todo Queue}. An additional TTS thread (\eg, F5-TTS~\cite{chen-etal-2024-f5tts}, MeloTTS~\cite{zhao2024melo}) will convert the text from the \emph{TTS Todo Queue} into audio and send it to the frontend.
\begin{table*}[t]
\caption{\textbf{Evaluation results on ASR tasks}: "CN" refers to Chinese speech, while "ENG" refers to English speech. The performance is measured using WER $\downarrow$ (Word Error Rate).}\label{tab:audio-performance}
\centering
\resizebox{0.9\linewidth}{!}{%
\begin{tabular}{lclccccccccc}
\toprule
\multirow{2}{*}{Method}& \textbf{} & \multirow{2}{*}{LLM}& \textbf{} & \multicolumn{2}{c}{{Wenetspeech (CN)}} & \textbf{} & \multicolumn{4}{c}{{Librispeech (ENG)}} \\
\cmidrule{5-6}\cmidrule{8-11}
 && && {Test\_Net $\downarrow$} & {Test\_Meeting $\downarrow$} & \textbf{} & {Dev\_clean $\downarrow$} & {Dev\_other $\downarrow$} & {Test\_clean$\downarrow$} & {Test\_other$\downarrow$} \\ 
\cmidrule{1-1}\cmidrule{3-3}\cmidrule{5-6}\cmidrule{8-11}
Qwen2-Audio \cite{chu2024qwen2} && Qwen2-7B \cite{yang2024qwen2}     && 7.8   & 8.4   &  & 1.3  & 3.4   & 1.6  & 3.6  \\
Mini-Omni \cite{xie2024mini}   && Qwen2-0.5B \cite{yang2024qwen2}   && -     & -     &  & 4.5  & 9.7   & 4.6  & 9.2  \\
VITA \cite{fu2024vitaopensourceinteractiveomni}        && Mixtral-8x7B \cite{jiang2024mixtral} && 12.2  & 16.5  &  & 7.6  & 16.6  & 8.1  & 18.4 \\
\rowcolor[HTML]{F2F3F5} IXC2.5-OL   && Qwen2-1.5B \cite{yang2024qwen2}   && 9.0   & 9.2   &  & 2.5  & 5.7   & 2.6  & 5.8  \\
\bottomrule
\end{tabular}%
}
\end{table*} 

\begin{table*}
    \centering
    \caption{\textbf{Evaluation results on MLVU benchmark.} IXC2.5-OL has demonstrated excellent performance, surpassing both open-source models and closed-source APIs, achieving SOTA at the 7B model scale.}
    \vspace{-0.5em}
    \footnotesize
    \begin{tabular}{lc|ccccccc|c}
    \toprule
    Method & Params & Topic Rea. & Anomaly Recog. & Needle QA & Ego Rea. & Plot QA & Action Or. & Action Co. & M-Avg\\
    \midrule
    \multicolumn{10}{l}{\textcolor{gray}{\emph{Closed-source APIs.}}} \\
    \midrule
    Claude-3-Opus & - & 67.2  & 43.5 & 21.6 & 40.2 & 47.8 & 18.2 & 16.7 & 36.5 \\
    Qwen-VL-Max & - & 67.4 & 63.5 & 40.3 & 40.9 & 43.3 & 25.0 & 14.8 & 42.2 \\
    GPT-4 Turbo & - & 79.5 & 68.0 & 45.9 &  47.4 & 60.6 & 26.5 &  16.1 & 49.2 \\
    GPT-4o & - & 87.4 & 74.5 & 64.8 &  57.1 & 65.1 & 56.7 &  46.3 & 64.6 \\
    \midrule
    \multicolumn{10}{l}{\textcolor{gray}{\emph{Open-source models.}}} \\
    \midrule
    MovieChat~\cite{song2024moviechat} & 7B & 29.5 & 25.0 & 24.2 & 24.7 & 25.8 & 28.6 & 22.8 & 25.8 \\
    LLaMA-VID~\cite{li2025llama} & 7B & 50.8 & 34.5 & 30.1 & 32.7 & 32.5 & 23.9 & 27.8 & 33.2 \\
    LLaVA-1.6~\cite{liu2024visual} & 7B & 60.6 & 41.0 & 43.1 & 38.4 & 41.0 & 25.5 & 25.7 & 39.3 \\
    ShareGPT4Video~\cite{chen2024sharegpt4video} & 7B & 75.8 & 51.5 & 47.6 & 43.2 & 48.4 & 34.0 & 23.3 & 46.4 \\
    VideoLlaMA2~\cite{cheng2024videollama} & 7B & 74.6 & 64.5 & 49.9 & 43.8 & 45.1 & 34.0 & 27.4 & 48.5 \\
    LongVA~\cite{zhang2024longva} & 7B & 83.3 & 58.5 & 69.3 & 50.0 & 67.2 &  38.6 & 27.2 & 56.3 \\
    IXC2.5~\cite{internlmxcomposer2_5} & 7B & - & - & - & - & - & - & - & 58.8 \\
    InternVL2~\cite{chen2024fargpt4vclosinggap} & 8B & - & - & - & - & - & - & - & 64.0 \\
    LLaVA-OneVision~\cite{li2024llava}  & 7B & - & - & - & - & - & - & - & 64.7 \\
    Video-XL~\cite{shu2024video}  & 7B & - & - & - & - & - & - & - & 64.9 \\
    \midrule
    \rowcolor[HTML]{F2F3F5} IXC2.5-OL & 7B & 84.1 & 68.5 & 76.6 & 60.8 & 75.1 & 57.1 & 41.3 & 66.2 \\
    \bottomrule
    \end{tabular}
    \label{tab:mlvu}
\end{table*}

\section{Experiments}
In this section, we validate the benchmark performance of our InternLM-XComposer2.5-OmniLive (IXC2.5-OL), including both audio and video benchmarks.

\begin{table}
    \centering
    \vspace{-1.0em}
    \footnotesize
    \setlength\tabcolsep{3.2pt}
    \caption{\textbf{Evaluation results on Video-MME benchmark.} IXC2.5-OL demonstrates performance close to that of the open-source SOTA.}
    \vspace{-0.5em}
    \begin{tabular}{lc|ccc|c}
    \toprule
    Method & Params & Short & Medium & Long & Overall \\
    \midrule
    \multicolumn{6}{l}{\textcolor{gray}{\emph{Closed-source APIs.}}} \\
    \midrule
    GPT-4V & - & 70.5 & 55.8 & 53.5 & 59.9 \\
    Claude 3.5 Sonnet & - & 71.0 & 57.4 & 51.2 & 60.0 \\
    GPT-4o mini & - & 72.5 & 63.1 & 58.6 & 64.8 \\
    GPT-4o & - & 80.0 & 70.3 & 65.3 &  71.9 \\
    Gemini 1.5 Pro & - & 81.7 & 74.3 & 67.4 &  75.0 \\
    \midrule
    \multicolumn{6}{l}{\textcolor{gray}{\emph{Open-source models.}}} \\
    \midrule
    ShareGPT4Video~\cite{chen2024sharegpt4video} & 7B & 48.3 & 36.3 & 35.0 & 39.9 \\
    VideoLlaMA2~\cite{cheng2024videollama} & 7B & - & - & - & 47.9 \\
    LongVA~\cite{zhang2024longva} & 7B & 61.1 & 50.4 & 46.2 & 52.6 \\
    Video-XL~\cite{shu2024video}  & 7B & 64.0 & 53.2 & 49.2 & 55.5 \\
    VITA~\cite{fu2024vitaopensourceinteractiveomni} & 8$\times$7B & 65.9 & 52.9 & 48.6 & 55.8 \\
    IXC2.5~\cite{internlmxcomposer2_5} & 7B & - & - & - & 55.8 \\
    InternVL2~\cite{chen2024fargpt4vclosinggap} & 8B & - & - & - &  56.3 \\
    LLaVA-OneVision~\cite{li2024llava}  & 7B & - & - & - & 58.2 \\
    mPLUG-Owl3~\cite{ye2024mplug}  & 7B & 70.0 & 57.7 & 50.1 & 59.3 \\
    MiniCPM-V 2.6~\cite{yao2024minicpm}  & 8B & - & - & - & 60.9 \\
    \midrule
    \rowcolor[HTML]{F2F3F5} IXC2.5-OL & 7B & 72.7 & 58.2 & 50.8 & 60.6 \\
    \bottomrule
    \end{tabular}
    \label{tab:videomme}
    \vspace{-10pt}
\end{table}

\subsection{Audio Benchmarks}

We evaluate our audio models on two prominent automatic speech recognition (ASR) benchmarks: Wenetspeech~\cite{zhang2022wenetspeech} for Chinese (CN) and LibriSpeech~\cite{panayotov2015librispeech} for English (EN). WenetSpeech includes two test sets: Test\_Net, which represents high-quality and relatively clean Chinese speech, and Test\_Meeting, which captures more challenging conversational scenarios. LibriSpeech consists of four splits: Dev\_clean and Test\_clean, which contain clean, high-quality English speech, and Dev\_other and Test\_other, which include noisier, more complex utterances.

As shown in Table \ref{tab:audio-performance}, our IXC2.5-OL demonstrates superior performance compared to recent streaming audio LLMs such as VITA and Mini-Omni, particularly achieving lower Word Error Rates (WER) across both CN and EN benchmarks with merely a lightweight 1.5B LLM.

\subsection{Video Benchmarks}
In Tables \ref{tab:mlvu}, \ref{tab:videomme}, \ref{tab:mmbench-video} and \ref{tab:mvbench}, we compare IXC2.5-OL with both closed-source APIs and open-source models on conventional video understanding benchmarks, including MLVU \cite{zhou2024mlvu}, Video-MME \cite{fu2024video}, MMBench-Video \cite{fang2024mmbench} and MVBench \cite{li2024mvbench}. Furthermore, we also assess the performance of different models on the recently proposed StreamingBench \cite{lin2024streamingbench}, which is designed to better evaluate performance for real-time video interactions. The results of this comparison are presented in Table \ref{tab:streamingbench}. For the video benchmarks, the base model utilizes 64 sampled frames for each video during evaluation.

\begin{table*}[t]
\centering
\caption{\textbf{Evaluation results on StreamingBench} for Real-Time Visual Understanding. Metrics include Object Perception (OP), Causal Reasoning (CR), Clips Summarization (CS), Attribute Perception (ATP), Event Understanding (EU), Text-Rich Understanding (TR), Prospective Reasoning (PR), Spatial Understanding (SU), Action Perception (ACP), and Counting (CT). IXC2.5-OL excels among all open-source models, and falling just short of the closed-source API, Gemini 1.5 Pro.}
\renewcommand{\arraystretch}{1}
\small
\setlength{\tabcolsep}{4pt}
\begin{tabular}{lc|cccccccccc|c}
\toprule
\multirow{2}{*}{Method} & \multirow{2}{*}{Params} & \multicolumn{11}{c}{Real-Time Visual Understanding} \\
\cmidrule{3-13}
& & OP & CR & CS & ATP & EU & TR & PR & SU & ACP & CT & Overall \\
\midrule
Human & -  & 89.47 & 92.00 & 93.60 & 91.47 & 95.65 & 92.52 & 88.00 & 88.75 & 89.74 & 91.30 & 91.46 \\
\midrule
\multicolumn{13}{l}{\textcolor{gray}{\emph{Closed-source APIs.}}} \\
\midrule
Claude 3.5 Sonnet & - & 80.49 & 77.34 & 82.02 & 81.73 & 72.33 & 75.39 & 61.11 & 61.79 & 69.32 & 43.09 & 72.44 \\
GPT-4o & - & 77.11 & 80.47 & 83.91 & 76.47 & 70.19 & 83.80 & 66.67 & 62.19 & 69.12 & 49.22 & 73.28 \\
Gemini 1.5 Pro & - & 79.02 & 80.47 & 83.54 & 79.67 & 80.00 & 84.74 & 77.78 & 64.23 & 71.95 & 48.70 & 75.69 \\
\midrule
\multicolumn{13}{l}{\textcolor{gray}{\emph{Open-source models.}}} \\
\midrule
VideoLLM-online~\cite{chen2024videollmonlineonlinevideolarge} & 8B  &  39.07 &40.06 &34.49& 31.05& 45.96 &32.40 &31.48& 34.16 &42.49 &27.89& 35.99 \\
VideoLLaMA2~\cite{cheng2024videollama} & 7B  & 55.86 & 55.47 & 57.41 & 58.17 & 52.80 & 43.61 & 39.21 & 42.68 &  45.61 & 35.23 & 49.52 \\
VILA-1.5~\cite{lin2024vilapretrainingvisuallanguage} & 8B & 53.68 & 49.22 & 70.98 & 56.86 & 53.42 & 53.89 & 54.63 & 48.78 & 50.14 & 17.62 & 52.32 \\
LongVA~\cite{zhang2024longva} & 7B  & 70.03 & 63.28 & 61.20 & 70.92 & 62.73 & 59.50 & 61.11 & 53.66 & 54.67 & 34.72 & 59.96 \\
InternVL2~\cite{chen2024fargpt4vclosinggap} & 8B  & 68.12 & 60.94 & 69.40 & 77.12 & 67.70 & 62.93 & 59.26 & 53.25 & 54.96 & 56.48 & 63.72 \\
Kangaroo~\cite{liu2024kangaroo} & 7B  & 71.12 & 84.38 & 70.66 & 73.20 & 67.08 & 61.68 & 56.48 & 55.69 & 62.04 & 38.86 & 64.60 \\
MiniCPM-V 2.6~\cite{yao2024minicpm} & 8B  & 71.93 & 71.09 & 77.92 & 75.82 & 64.60 & 65.73 & 70.37 & 56.10  & 62.32 & 53.37 & 67.44 \\
Qwen2-VL~\cite{wang2024qwen2vlenhancingvisionlanguagemodels} & 7B  & 75.20 & 82.81 & 73.19 & 77.45 & 68.32 & 71.03 & 72.22 & 61.19 & 69.04 & 46.11 & 69.04 \\
LLaVA-OneVision~\cite{li2024llava} & 7B  & 80.38 & 74.22 & 76.03 & 80.72 & 72.67 & 71.65 & 67.59 & 65.45 & 65.72 & 45.08 & 71.12 \\
\midrule
\rowcolor[HTML]{F2F3F5} IXC2.5-OL  & 7B  & 82.83 & 73.77 & 78.66 & 82.95 & 72.50 & 76.01 & 61.11 & 60.67 & 71.59 & 58.85 & 73.79 \\
\bottomrule
\end{tabular}
\label{tab:streamingbench}
\end{table*}

\begin{table*}
\centering
\footnotesize
\caption{\textbf{Evaluation results on MMBench-Video.} Tasks include Coarse Perception (CP), Single-Instance Finegrained Perception (FP-S), Cross-Instance Finegrained Perception (FP-C), Hallucination (HL), Logic Reasoning (LR), Attribute Reasoning (AR), Relation Reasoning (RR), Commonsense Reasoning (CSR), and Temporal Reasoning (TP).}
\renewcommand{\arraystretch}{1.3}
\setlength{\tabcolsep}{4pt}
\begin{tabular}{lc|cccc|c|ccccc|c|c}
\toprule
\multirow{2}{*}{Method} & \multirow{2}{*}{Params} & \multicolumn{5}{c|}{Perception} & \multicolumn{6}{c|}{Reasoning} & \multirow{2}{*}{Overall} \\
\cmidrule{3-13}
& & CP & FP-S & FP-C & HL & Mean & LR & AR & RR & CSR & TP & Mean &  \\
\midrule
\multicolumn{13}{l}{\textcolor{gray}{\emph{Closed-source APIs.}}} \\
\midrule
Claude 3.5 Sonnet & - & 1.57 & 1.39 & 1.07 & 1.40 & 1.38 & 1.13 & 1.70 & 1.48 & 1.54 & 1.04 & 1.35 & 1.38 \\
Gemini 1.0 Pro & - & 1.61 & 1.56 & 1.30 & 0.65 & 1.50 & 1.15 & 1.57 & 1.55 & 1.36 & 1.33 & 1.39 & 1.48 \\
Gemini 1.5 Pro & - & 1.99 & 2.04 & 1.70 & 1.90 & 1.98 & 1.98 & 2.02 & 1.92 & 1.78 & 1.63 & 1.86 & 1.94 \\
GPT-4V & - & 1.83 & 1.65 & 1.40 & 1.76 & 1.66 & 1.45 & 1.91 & 1.86 & 1.83 & 1.53 & 1.69 & 1.68 \\
GPT-4o & - & 2.23 & 2.24 & 2.01 & 1.90 & 2.19 & 2.11 & 2.12 & 2.17 & 1.94 & 1.97 & 2.08 & 2.15 \\
\midrule
\multicolumn{13}{l}{\textcolor{gray}{\emph{Open-source models.}}} \\
\midrule
MovieLLM~\cite{song2024moviellm} & 7B  &  0.95 &0.82  &0.70 & 0.15& 0.81 &0.52 &1.12 & 1.22 &0.54 &1.05& 0.97 & 0.87 \\
LLaVA-OneVision~\cite{li2024llava} & 72B  &  1.22 &1.07  &0.90 & 0.21& 1.03 &0.76 &0.96 & 0.55 &0.81 &0.48& 0.70 & 0.94 \\
PLLaVA~\cite{xu2024pllavaparameterfreellava} & 7B  &  1.08 &1.06  &0.86 & 0.52 & 1.02 &0.64 &1.25 & 1.17 &0.98  &1.01& 1.03 & 1.03 \\
ShareGPT4Video~\cite{chen2024sharegpt4video} & 7B  &  1.20 &1.05 &1.00 & 0.32 & 1.04 &0.89 &1.06 & 1.19 &1.01&0.99& 1.03 & 1.05 \\
VideoStreaming~\cite{qian2024streaminglongvideounderstanding} & 7B  &  1.38 &1.13 &0.8 & 0.32 & 1.13 &0.77 &1.27 & 1.11 &1.01&1.10& 1.09 & 1.12 \\
LLaVA-NeXT-Video~\cite{zhang2024llavanextvideo} & 7B  &  1.35 &1.15 &0.97 & 0.58 & 1.14 &0.64 &1.38 & 1.30 &1.27&1.03& 1.13 & 1.14 \\
VILA1.5~\cite{lin2024vilapretrainingvisuallanguage} & 13B  & 1.51 &1.45 &1.26 & 0.24 & 1.39 &0.80 & 1.52 & 1.30 &1.40 &1.28& 1.28 & 1.36 \\
InternVL2~\cite{chen2024fargpt4vclosinggap} & 8B  & 1.41 &1.37 &1.15 & 0.19 & 1.30 &0.90 & 1.34 & 1.38 &1.14 &1.00& 1.16 & 1.26 \\
Qwen2-VL~\cite{wang2024qwen2vlenhancingvisionlanguagemodels} & 7B  & 1.63 &1.51 &1.19 & 0.55 &1.46 & 1.16 &1.56 & 1.49 & 1.37 &1.21& 1.35 & 1.44 \\
\midrule
\rowcolor[HTML]{F2F3F5} IXC2.5-OL & 7B  & 1.53 &1.61 &1.20 & 0.15 &1.49 & 0.93 &1.44 & 1.57 & 1.30 &1.08& 1.25 & 1.42 \\
\bottomrule
\end{tabular}
\label{tab:mmbench-video}
\end{table*}

\begin{table*}[t]
    \centering
    \caption{\textbf{Evaluatation results on MVBench.} Tasks include Action Sequence (AS), Action Prediction (AP), Action Antonym (AA), Fine-grained Action (FA), Unexpected Action (UA), Object Existence (OE), Object Interaction (OI), Object Shuffle (OS), Moving Direction (MD), Action Localization (AL), Scene Transition (ST), Action Count (AC), Moving Count (MC), Moving Attribute (MA), State Change (SC), Fine-grained Pose (FP), Character Order (CO), Egocentric Navigation (EN), Episodic Reasoning (ER), and Counterfactual Inference (CI).}
    \footnotesize
    \setlength\tabcolsep{2.4pt}
    \begin{tabular}{lc|cccccccccccccccccccc|c}
    \toprule
    Method & Params & AS & AP & AA & FA & UA & OE & OI & OS & MD& AL& ST& AC& MC& MA& SC& FP& CO& EN& ER& CI& Avg \\
    \midrule
    \multicolumn{10}{l}{\textcolor{gray}{\emph{Closed-source APIs.}}} \\
    \midrule
    GPT-4V & - & 55.5& 63.5& 72.0& 46.5& 73.5& 18.5& 59.0& 29.5& 12.0 &40.5& 83.5& 39.0 &12.0& 22.5& 45.0& 47.5& 52.0& 31.0& 59.0& 11.0&43.5 \\
    GPT-4o & - & 61.5 &56.5&72.0&54.0&82.0&62.5&66.5&44.0&36.5&33.5&93.0&54.5&33.5&54.5&53.5&74.5&71.5&32.5&71.0&42.5&57.5 \\
    \midrule
    \multicolumn{10}{l}{\textcolor{gray}{\emph{Open-source models.}}} \\
    \midrule
    VideoLLaMA~\cite{zhang2023video} & 7B & 27.5 &25.5& 51.0& 29.0& 39.0& 48.0& 40.5& 38.0& 22.5& 22.5& 43.0& 34.0& 22.5& 32.5& 45.5& 32.5& 40.0& 30.0& 21.0& 37.0&34.1 \\
    VideoChat~\cite{li2023videochat} & 7B &33.5 & 26.5& 56.0& 33.5& 40.5& 53.0& 40.5& 30.0& 25.5& 27.0& 48.5& 35.0& 20.5& 42.5& 46.0& 26.5& 41.0& 23.5& 23.5& 36.0&35.5\\
    MiniCPM-V 2.6~\cite{yao2024minicpm} & 7B & 38.0&43.0&63.0&35.5&67.5&55.5&46.0&35.5&25.5&33.0&77.5&48.0&37.0&54.0&42.5&40.0&31.0&38.0&43.0&40.5&44.7 \\
    VideoChat2~\cite{li2024mvbench} & 7B & 66.0& 47.5& 83.5& 49.5& 60.0 &58.0& 71.5& 42.5& 23.0& 23.0& 88.5& 39.0& 42.0& 58.5& 44.0& 49.0& 36.5& 35.0& 40.5& 65.5&51.1 \\
    Qwen2-VL~\cite{wang2024qwen2vlenhancingvisionlanguagemodels} & 7B  & 51.0&58.0&77.5&47.0&64.0&63.0&65.5&40.0&25.5&35.5&77.0&43.5&47.0&62.0&42.0&61.5&49.5&41.5&47.5&41.5&52.0\\
    PLLaVA~\cite{xu2024pllavaparameterfreellava} & 34B  & 65.0&53.0&83.5&45.0&77.5&70.0&64.5&38.5&37.5&49.0&89.5&41.5&43.5&70.0&53.0&52.5&65.0&39.5&60.5&58.0&57.8\\
    LLaVA-OneVision~\cite{li2024llava} & 72B & 63.0&58.0&84.5&46.5&85.5&64.0&73.5&41.5&37.0&69.0&95.0&47.5&47.5&75.5&53.5&52.0&70.5&34.0&64.0&54.5&60.8\\
    InternVL2~\cite{chen2024fargpt4vclosinggap} & 8B & 75.0 &62.0&83.5&40.5&69.5&96.0&72.0&29.5&58.0&53.0&88.5&39.5&83.0&97.0&51.0&78.5&65.0&33.0&48.0&67.0&64.5 \\
    \midrule
    \rowcolor[HTML]{F2F3F5} IXC2.5-OL & 7B  & 84.5& 81.0& 75.0& 46.0& 81.0& 92.0& 79.5& 36.5& 83.0& 47.0& 90.0& 60.5& 75.0, & 93.0& 58.0& 60.5& 74.0& 42.0& 53.0& 62.0 & 68.7 \\
    \bottomrule
    \end{tabular}
    \label{tab:mvbench}
\end{table*}

\noindent \paragraph{MLVU} 
MLVU is a comprehensive benchmark designed for evaluating Multimodal Large Language Models in Long Video Understanding tasks. The videos range from 3 minutes to 2 hours and include nine distinct evaluation tasks. Here, we evaluate seven multi-choice tasks, including Topic Reasoning, Anomaly Recognition, Needle QA, Ego Reasoning, Plot QA, Action Order, and Action Count. The detailed comparisons are given in Table \ref{tab:mlvu}. The IXC2.5-OL exhibits state-of-the-art (SOTA) performance among closed-source APIs, and open-source models with parameters less than 10 billion, surpassing the previous SOTA by $1.3\%$ for Video-XL, $1.6\%$ for GPT-4o.

\noindent \paragraph{Video-MME}
Video-MME is a high-quality video benchmark. The videos are collected from 6 primary visual domains with 30 subfields to ensure broad scenario generalizability, encompassing both short-, medium-, and long-term videos, ranging from 11 seconds to 1 hour. As demonstrated in Table \ref{tab:videomme}, the IXC2.5-OL exhibits competitive performance on this benchmark, comparable to previous SOTA MiniCPM-V 2.6.

\noindent \paragraph{StreamingBench}
StreamingBench is a streaming video benchmark designed for real-time video evaluation. It comprises 18 tasks, showcasing 900 videos and 4,500 human-curated QA pairs. In this context, we focus on assessing visual understanding in real-time. Table \ref{tab:streamingbench} illustrates the comparative analysis, demonstrating that IXC2.5-OL excels among all open-source models, achieving a $2.67\%$ improvement over the previous state-of-the-art model, LLaVA-OneVision, and falling just short of the closed-source API, Gemini 1.5 Pro. This performance solidifies IXC2.5-OL's remarkable prowess in real-time video interaction.

\noindent \paragraph{MMBench-Video}
MMBench-Video is a free-form QA video benchmark consisting of 600 videos and 2000 QA pairs. The duration of each video varies from 30 seconds to 6 minutes. Given the open-ended nature of the answers, the benchmark utilizes GPT-4-based evaluation to enhance quality in terms of accuracy, consistency, and alignment with human judgment. The results are presented in Table \ref{tab:mmbench-video}. IXC2.5-OL demonstrates state-of-the-art performance on perception tasks and comparable performance on overall evaluations.

\noindent \paragraph{MVBench}
MVBench is a video benchmark that emphasizes temporal understanding. It encompasses 20 challenging video tasks that cannot be effectively addressed using a single frame. As shown in Table \ref{tab:mvbench}, IXC2.5-OL, despite having a smaller 7B parameter size, has outperformed both the GPT-4 series and the 72B open-source model LLaVA-OneVision, demonstrating its strong capability in understanding video temporal dynamics.

\section{Conclusion}
We have presented IXC2.5-OL, a real-time streaming model that advances multi-modal text, audio, and visual capabilities with long-term memory.
IXC2.5-OL empowers users to engage in dynamic and interactive experiences.
Our model's real-time processing enables fluid and responsive interactions, allowing users to engage with ever-changing environments of multimodal data seamlessly, providing a more intuitive and efficient user experience.
Our future work will focus on reducing system latency to provide a seamless user experience.
\clearpage
{
    \small
    \bibliographystyle{ieeenat_fullname}
    \bibliography{main}
}


\end{document}